\documentclass[runningheads]{llncs}
\usepackage[T1]{fontenc}
\usepackage{cite}
\usepackage{enumitem}
\usepackage{booktabs}
\usepackage{multirow}
\usepackage{amssymb}    
\usepackage{float} 
\usepackage{xcolor}
\usepackage{pifont}
\usepackage{amsmath}
\usepackage{bbm}
\usepackage{microtype}
\usepackage[colorlinks=true,linkcolor=blue,citecolor=blue,urlcolor=blue]{hyperref}
\usepackage{CJKutf8}
\usepackage{graphicx}
\begin{document}
\title{LaiDA: Linguistics-aware In-context Learning with Data Augmentation for Metaphor Components Identification}
\titlerunning{Linguistics-aware In-context Learning with Data Augmentation for MCI}
% If the paper title is too long for the running head, you can set
% an abbreviated paper title here
%
\author{Hongde Liu \and
Chenyuan He \and
Feiyang Meng \and Changyong Niu \and Yuxiang Jia\thanks{Corresponding author}}

\authorrunning{H. Liu et al.}
% First names are abbreviated in the running head.
% If there are more than two authors, 'et al.' is used.

\institute{School of Computer and Artificial Intelligence, Zhengzhou University, Zhengzhou, China\\
\email{\{lhd\_1013,hechenyuan\_nlp\}@gs.zzu.edu.cn}
\\
\email{\{3492298894\}@qq.com} 
\\
\email{\{iecyniu,ieyxjia\}@zzu.edu.cn}}
\maketitle              % typeset the header of the contribution
\begin{abstract}
Metaphor Components Identification (MCI) contributes to enhancing machine understanding of metaphors, thereby advancing downstream natural language processing tasks. However, the complexity, diversity, and dependency on context and background knowledge pose significant challenges for MCI. Large language models (LLMs) offer new avenues for accurate comprehension of complex natural language texts due to their strong semantic analysis and extensive commonsense knowledge. In this research, a new LLM-based framework is proposed, named \textbf{L}inguistics-\textbf{a}ware \textbf{I}n-context Learning with \textbf{D}ata \textbf{A}ugmentation (LaiDA). Specifically, ChatGPT and supervised fine-tuning are utilized to tailor a high-quality dataset. LaiDA incorporates a simile dataset for pre-training. A graph attention network encoder generates linguistically rich feature representations to retrieve similar examples. Subsequently, LLM is fine-tuned with prompts that integrate linguistically similar examples. LaiDA ranked $2^{nd}$ in Subtask 2 of NLPCC2024 Shared Task 9, demonstrating its effectiveness. Code and data are available at \url{https://github.com/WXLJZ/LaiDA}.

\end{abstract}
\section{Introduction}
\begin{figure*}[!h]
    \centering
    \includegraphics[width=\textwidth]{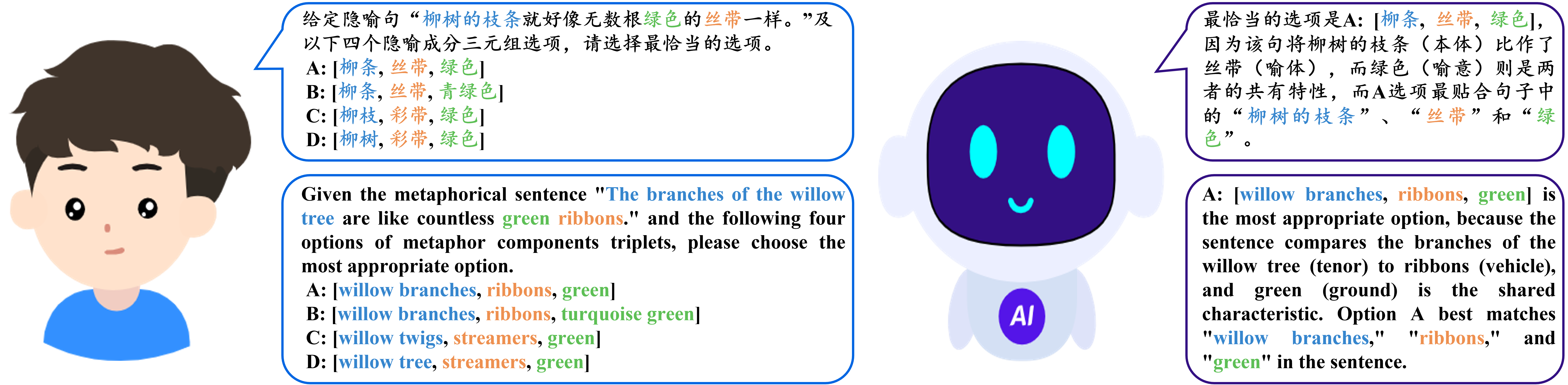}
    \caption{An example of the metaphor components identification.}
    \label{fig:fig3}
\end{figure*}

As a significant phenomenon in natural language, metaphors are prevalent in daily communication, literary works, and scientific discourse. As illustrated in Fig.\ref{fig:fig3}, Metaphor Components Identification (MCI), which involves accurately identifying and interpreting the components of metaphors (typically including the tenor, vehicle, and ground), holds substantial value for various applications in natural language processing (NLP), such as sentiment analysis, information retrieval, and machine translation \cite{ge2023survey,mao2018word,socher2013recursive}. However, the complexity and diversity of metaphors, along with their reliance on context and background knowledge, present significant challenges.

Early research on MCI relied on manually crafted rules and dictionaries, which were limited in scale and flexibility \cite{ge2023survey,shutova2010metaphor,krishnakumaran2007hunting}. The advent of machine learning, especially deep learning, shifted the focus to neural network models based on word embeddings and sequence models \cite{mao2018word,liu2018neural}, enhancing recognition capabilities. Nonetheless, these methods still face challenges in contextual understanding and model generalization.

The recent rise of large language models (LLMs) has advanced MCI. Through extensive pre-training and fine-tuning, LLMs can effectively tackle complex and variable metaphors and demonstrate remarkable generalization capabilities. The emerging in-context learning (ICL) enhances task performance by incorporating concise, task-relevant instructions, demonstrations, and examples in prompts \cite{dong2022survey}. Notably, metaphors, as a unique literary form, exhibit distinctive linguistic features, and current research \cite{liu2022makes,yang2024faima} indicates that integrating linguistically similar examples into fine-tuning instructions further improves LLM performance on specific tasks.

In light of the above, we propose a framework that integrates linguistics-aware in-context learning and data augmentation, named LaiDA, to enhance the precision and efficiency of MCI. Specifically, LaiDA leverages ChatGPT to construct a small, high-quality benchmark dataset, which is used to fine-tune an LLM to generate a larger dataset. Additionally, to grasp fundamental metaphorical patterns, LaiDA employs a pre-training dataset of similes for data augmentation. For the in-context learning module, LaiDA retrieves linguistically similar examples from the training set based on linguistic feature representations encoded by a graph attention network encoder, inspired by FaiMA \cite{yang2024faima}. Finally, based on the constructed training set, we use prompts incorporating linguistically similar examples for fine-tuning. Experimental results and analysis demonstrate the effectiveness of LaiDA in MCI. Our main contributions are summarized as follows:

\begin{itemize}[label=$\bullet$, topsep=0pt, partopsep=0pt, parsep=0pt, itemsep=0pt]
    \item We propose an LLM-based framework (LaiDA), integrating linguistics-aware in-context learning and data augmentation, capable of effectively recognizing metaphor components.
    \item In the linguistics-aware in-context learning module, we propose a sentence encoder model that captures linguistic features in metaphors, and design a retriever to search for linguistically similar examples within large datasets based on the representations encoded.
    \item A metaphor dataset is constructed automatically for MCI, leveraging ChatGPT to generate initial data, fine-tuning a smaller LLM, and applying the fine-tuned model to process the remaining data, thereby significantly reducing labor and time costs in data annotation.
\end{itemize}

\section{Related Work}
In recent years, the emergence of pre-trained models has provided new solutions to many challenges in the field of artificial intelligence, and the metaphor computing community has also been greatly inspired.

\textbf{Metaphor identification} has always been a hot research topic in the community, as the foundation of computational metaphor processing. Classical Metaphor Theory, Conceptual Metaphor Theory \cite{lakoff2008metaphors}, and Selectional Preference Violation Theory \cite{wilks1978making} serve as the theoretical foundations for many subsequent metaphor identification methods. Song et al. \cite{song2021verb} proposed MrBERT for verb metaphor detection, extracting and representing context from different perspectives, and modeling the relationship between the target verb and its context. Mao and Li\cite{mao2021bridging} adopted a multi-task learning approach, treating metaphor identification as the primary task and part-of-speech tagging as an auxiliary task. They employed a novel gating mechanism for the bridging of multi-task learning towers that demonstrated excellent performance in both tasks.

\textbf{Metaphor interpretation}, as a follow-up to metaphor identification, has also evolved. Bizzoni and Lappin \cite{bizzoni2018predicting} proposed a CNN-LSTM framework to capture the semantic representations of metaphors and literal expressions. Given a metaphorical sentence and a literal candidate, the model ranks the candidates, with the top-ranked candidate being considered the paraphrase of the metaphorical sentence. However, this model requires a large number of customized candidate paraphrases, which is inconvenient for practical applications. Song et al. \cite{song2020knowledge} proposed a knowledge graph embedding-based method, representing each specific metaphor as a metaphorical triplet (target, attribute, source), thereby treating metaphor interpretation as a knowledge graph completion task.

\textbf{In-context Learning} has demonstrated remarkable performance in pre-trained models, especially large language models. ICL is highly sensitive to the similarity and diversity of few-shot demonstrations \cite{xu2024context}. For example, Liu et al. \cite{liu2022makes} proposed KATE, which used KNN to compute the similarity distance between input queries and demonstrations. Yu et al. \cite{yugenerate} introduced a clustering-based prompting method to generate diverse demonstrations, increasing the likelihood of covering correct answers and greatly expanding demonstration diversity. Levy et al. \cite{levy2023diverse} proposed a method to maximize demonstration coverage from lexical or syntactic perspectives, outperforming similarity-based demonstration selection on three compositional generalization semantic parsing datasets.

\begin{figure*}[!t]
    \centering
    \includegraphics[width=\textwidth]{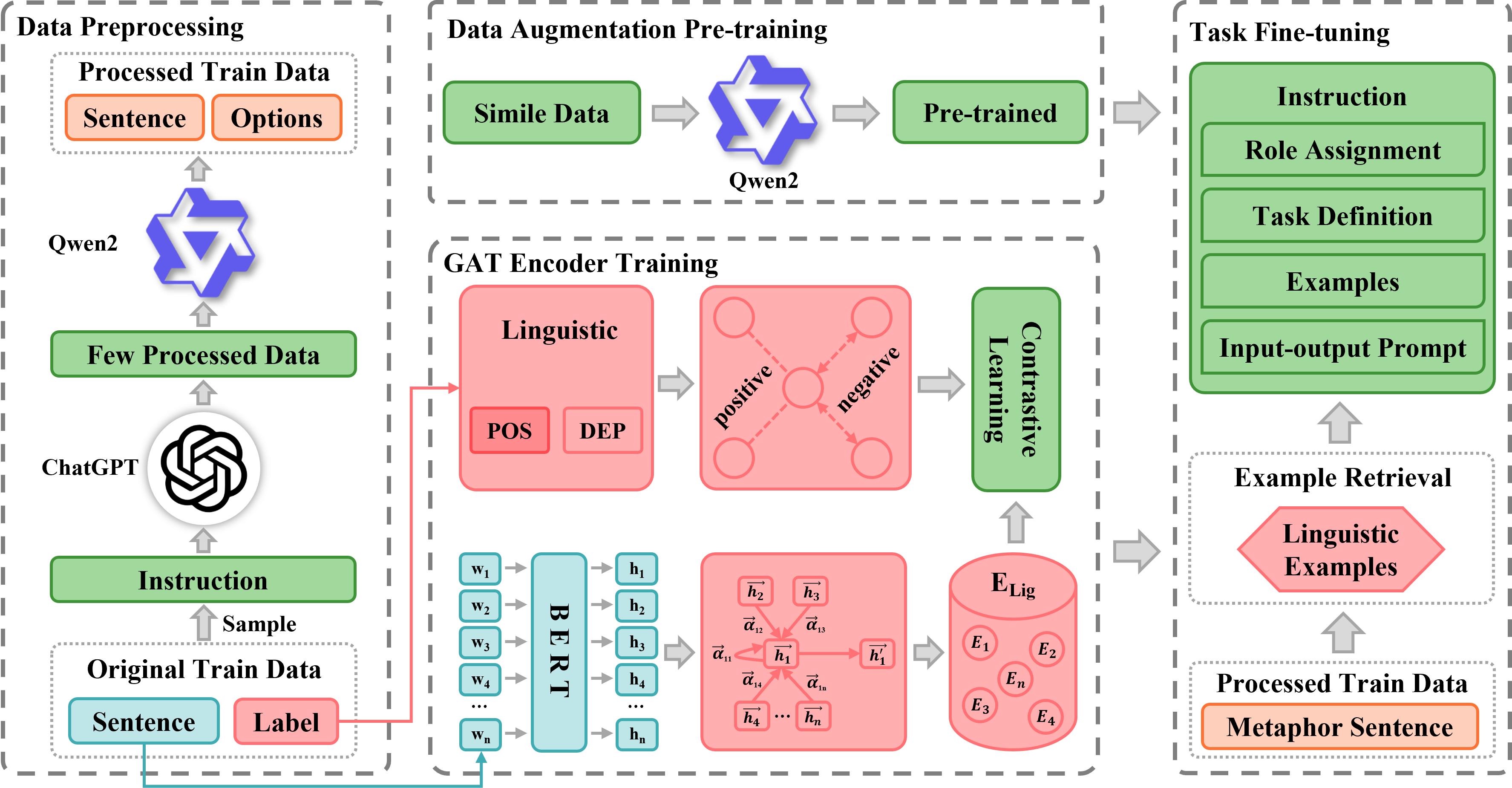}
    \caption{The training workflow of LaiDA.}
    \label{fig:fig1}
\end{figure*}

\section{Methodology}
In the following, we introduce our proposed LaiDA framework. As shown in Fig.\ref{fig:fig1}, the LaiDA framework consists of four modules: data preprocessing (\ref{part: data process}), data augmentation pre-training (\ref{part: pretraining}), graph attention network encoder (\ref{part: GAT encoder}), and task fine-tuning (\ref{part: task fine-tune}).

\subsection{Task Definition}
The metaphor components identification task aims to identify and interpret the components of a metaphor from a metaphorical sentence, which include the tenor $T$, vehicle $V$, and ground $G$, where the ground reflects the commonality between the tenor and vehicle. Given a metaphorical sentence $S$ and four metaphor component triplets $O = [T, V, G]$, $O \in \{A, B, C, D\}$, the objective of the task is to select the metaphor components triplet that best matches the given metaphorical sentence from the four candidates.

\subsection{Data Preprocessing}
\label{part: data process}
Data preprocessing is required to integrate these components into triplets $O = [T, V, G]$ and generate distractor triplets $O' = [T', V', G']$ with incorrectly identified metaphor components, as the original training set \cite{shao2024cmdag} only contains metaphorical sentences and their corresponding metaphor components. To optimize the data processing workflow and reduce economic and time costs, we ingeniously utilized data generated by a larger LLM to guide a smaller LLM.

\begin{figure*}[!t]
    \centering
    \includegraphics[width=\textwidth]{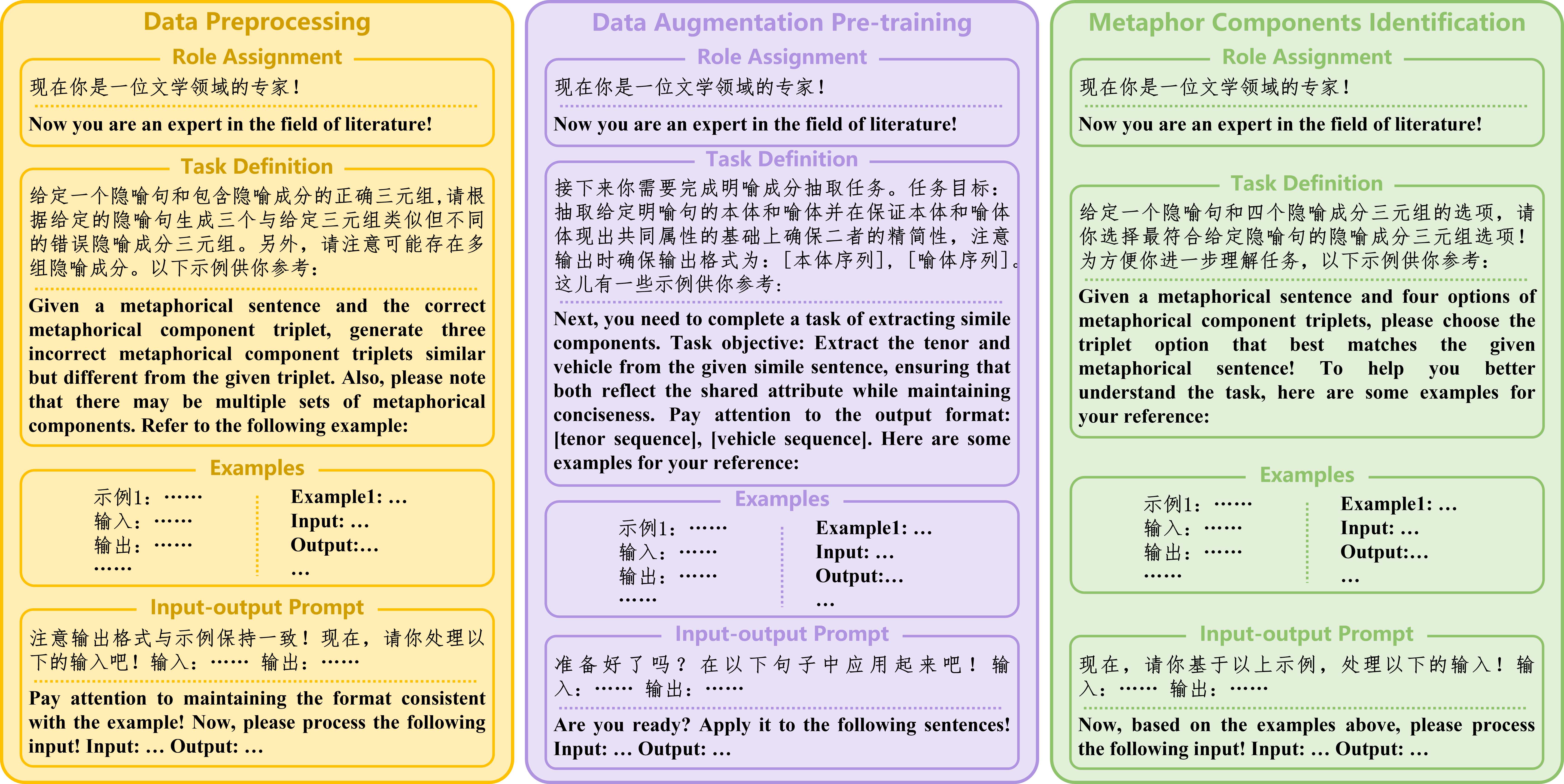}
    \caption{The prompts used during supervised fine-tuning.}
    \label{fig:fig2}
\end{figure*}

Guided by the prompt shown in the left figure of Fig.~\ref{fig:fig2}, ChatGPT\footnote{\url{https://chatgpt.com/}} is utilized to generate three distractor triplets $O'_1, O'_2, O'_3$ in a small sample dataset that is drawn from the curated original dataset. Subsequently, a dataset $\mathbbm{D}_{sampled}$ based on these processed samples is constructed, and a smaller LLM than ChatGPT is fine-tuned on $\mathbbm{D}_{sampled}$. The fine-tuned model is used to comprehensively process the remaining training data, organizing each datum into a metaphorical sentence with four metaphor component triplet options. Furthermore, randomization is implemented to shuffle the four options, ensuring the correct triplet is randomly distributed among the A, B, C, and D.

\subsection{Data Augmentation Pre-training}
\label{part: pretraining}
The data augmentation pre-training module is aimed at enhancing the model's ability to recognize metaphors by learning their features and patterns in advance. We employed the metaphor dataset by Liu et al. \cite{liu2018neural}, annotated with the tenors and vehicles of metaphorical sentences, to fine-tune the LLM under the guidance of the middle prompt depicted in Fig.\ref{fig:fig2}. While the dataset focuses on similes, a relatively straightforward form of metaphor, it provides valuable initial exposure to metaphors.

\subsection{Graph Attention Network Encoder}
\label{part: GAT encoder}
As shown in Fig.\ref{fig:fig1}, the GAT encoder can be subdivided into three units: the construction of positive and negative sample matrices for linguistic features, the graph attention network, and contrastive learning. Each of these units is described in detail below:
\subsubsection{Linguistic Feature Matrix Construction}
\label{part: Linguistic Feature Matrix}
We employ the following heuristic rules to construct positive and negative sample matrices for linguistic features as one of the inputs for the contrastive learning unit. Linguistic features encompass various aspects, here focusing on part-of-speech in lexical analysis and syntactic dependency relations in syntactic analysis as elements of linguistic features.

Given a set of metaphorical sentences $\mathbbm{S} = \{s_1, s_2, ..., s_n\}$, where $s_i$ represents the $i^{th}$ sentence in $\mathbbm{S}$, each sentence $s_i$ is parsed to build a feature matrix $F \in \mathbbm{R}^{l\times l\times 2}$ comprising its corresponding part-of-speech matrix $P_i$ and dependency relation matrix $D_i$. The central words $\mathbbm{C} = \{c_1, c_2, ..., c_m\}$ in $s_i$ are identified by selecting the word with the most non-zero syntactic dependency relations in $s_i$. 

The distance $L$ between each pair of sentences $(S_1, S_2)$ is calculated according to the following process. The Gaussian function is utilized to compute the distance between each word and the central word in a sentence. Next, the weighted Hamming distance between the two feature matrices is calculated, which is defined as follows:
\begin{equation}
    L(S_1, S_2) = \frac{1}{\left|\mathbbm{C}_1\right|\left|\mathbbm{C}_2\right|} \sum_{i=1}^{\left|\mathbbm{C}_1\right|}\sum_{j=i}^{\left|\mathbbm{C}_2\right|} \left[e^{-\frac{(1 - \textit{idx})^2}{2 \sigma^2}}, ...,e^{-\frac{(n - \textit{idx})^2}{2 \sigma^2}}\right] \circ H\left[F_1(i),F_2(j)\right]
\end{equation}
Here, $\circ$ denotes the dot product operation, with $idx$ indicating the position of the central word $c_i$ in $S_1$. $\mathbbm{C}_1$ and $\mathbbm{C}_2$ house the central words corresponding to $S_1$ and $S_2$, respectively. Meanwhile, $F_1$ and $F_2$ are the feature representations of $S_1$ and $S_2$. The Hamming distance, expressed as $\textit{H}$, takes the value 1 when the corresponding elements of the two vectors are equal and 0 otherwise.

The sigmoid function is used to transform the distance $L(S_1, S_2)$ into the final similarity score (0 to 1), where a higher score indicates greater semantic similarity. Then the top K\% elements of each row are considered as positive samples, while the rest are negative samples, to construct the final positive and negative sample matrix of linguistic features.

\subsubsection{Graph Attention Network}
\label{part: GAT}
GAT, with its dynamic attention mechanism, can adaptively assign weights based on the features of nodes and their neighboring nodes, thus capturing important feature information in the graph more accurately. Here, the tokens of sentence $S$ are regarded as nodes in the graph attention network to capture the linguistic features reflected in the sentence. The details are as follows:

The GAT layer takes the tokens $(w_1,w_2, ..., w_n)$ encoded by the pre-trained model BERT \cite{devlin2018bert} as input vectors $H = (h_1, h_2, ..., h_n)$. An adaptive adjacency matrix $A^{Lig} = Sigmoid(HW^{Lig}H^T)$ is used as the feature propagation matrix for these token vectors, where $W^{Lig}$ represents the learnable weights of the linguistic features.

The attention coefficients between nodes are computed as:
\begin{equation}
\alpha_{ij}=\frac{\exp(\text{LeakyReLU}(\vec{a}^\mathsf{T}[W_ah_i\|W_ah_j]))}{\sum_{A_{ij}^{Lig}>\delta}\exp(\text{LeakyReLU}(\vec{a}^\mathsf{T}[W_ah_i\|W_ah_k]))}
\end{equation}
where $.^\mathsf{T}$ represents transposition and $\|$ is the concatenation operation, $\delta$ serves as a threshold to filter out noise in the adjacency matrix, $\vec{a}$ and $W_a$ are learnable weights.

After that, token-level representations are computed through the attention mechanism:
\begin{equation}
  \label{eq:equation5}
  E_i = LayerNorm(\sum_{A_{ij}^{Lig}>\delta}\alpha_{ij}A_{ij}^{Lig}W_ah_j),
\end{equation}
where $i$ denotes the $i^{th}$ token in the sentence. However, to obtain the linguistic feature vector for the entire sentence, serving as another input for contrastive learning, integrating all token-level representations is required. Here, an average pooling operation is adopted, which is represented as $E^{Lig} = AVG(E_1, E_2, ..., E_n)$.

\subsubsection{Contrastive Learning}
The previous two units of the GAT encoder are connected through the contrastive learning unit during training. The GAT continuously optimizes the sentence-level representations $E^{Lig}$ based on the constructed positive and negative sample matrices of linguistic features, ultimately obtaining sentence-level representations rich in linguistic features. For any sentence $S$, its positive samples are denoted as $\mathcal{P}$, and its negative samples as $\mathcal{N}$. The contrastive learning loss is calculated as follows:
\begin{equation}
    \label{eq:equation8}    
    \mathcal{L}_{Lig} = \sum_{S_{i}}\left[-\log \frac{\sum_{S_{j} \in \mathcal{P}_{i}} \exp \left(sigmoid\left((E_{ij} \circ E_{ij}^{T}) / \tau \right)\right)}{\sum_{S_{k} \in \mathcal{N}_{i}} \exp \left(sigmoid\left((E_{ik} \circ E_{ik}^{T}) / \tau \right)\right)}\right],
\end{equation}
where, $\circ$ denotes dot product operation, $E$ represents the linguistic sentence-level representations, and $\tau$ is the temperature parameter. The sentence-level representations $E^{Lig}$ of the GAT are optimized by continuously minimizing the loss $\mathcal{L}_{Lig}$.

\subsection{Task Fine-tuning}
\label{part: task fine-tune}
In this section, all previous modules are integrated to fine-tune for MCI. The training data constructed from the data preprocessing module (\ref{part: data process}) is encoded by the GAT encoder (\ref{part: GAT encoder}) into sentence representation rich in linguistic features, and FAISS (Facebook AI Similarity Search)\footnote{\url{https://github.com/facebookresearch/faiss}} is used to perform approximate nearest neighbor search within the training set, based on these sentence vectors, to retrieve three linguistically similar examples. Subsequently, we followed the instructions that integrated the prompt (Fig.~\ref{fig:fig2}, right panel) and similar examples, using the pre-trained model obtained from the data augmentation pre-training module (\ref{part: pretraining}) as the base model to further fine-tune the LLM.

\section{Experiments}

\subsection{Datasets}
We conduct experiments using the dataset provided by NLPCC2024 Shared Task 9 \cite{qu2024overviewnlpcc2024shared} Subtask 2: Metaphor Components Identification\footnote{\url{https://github.com/xingweiqu/NLPCC-2024-Shared-Task-9}}, which is divided into training, validation, and test sets. The statistical details are shown in Table~\ref{tab:tab1}, where \textbf{Single} and \textbf{Multiple} refer to the proportion of metaphorical sentences containing single and multiple groups of metaphor components in the dataset, respectively. \textbf{Avg.Sub-sents} indicates the average number of sub-sentences in a single metaphorical sentence, with sub-sentences separated by commas or spaces. \textbf{Simile} indicates the proportion of simile sentences in the dataset. The training set is selected from the dataset proposed by Shao et al.\cite{shao2024cmdag} in a ratio of \textbf{Single:Multiple = 9:1}. The dataset originates from various highly refined literary forms, including poetry, prose, lyrics, etc., which are not conducive to machine understanding. Notably, illustrated in Table~\ref{tab:tab1}, the expression forms of metaphorical sentences in the validation and test sets differ from those in the training set, being more simplistic. Therefore, during data augmentation pre-training, we applied the Chinese simile corpus\footnote{\url{https://github.com/cnunlp/Chinese-Simile-Recognition-Dataset}} created by Liu et al.\cite{liu2018neural} to familiarize the model with simpler metaphors and to partially bridge the gap between the training and test sets.

\begin{table*}[t!]
    \caption{The statistics of MCI dataset. }
    \label{tab:tab1}
    \centering
    \begin{tabular}{c|c|c|c|c|c}
        \toprule[1.5pt]
        \textbf{Datasets} & \textbf{Sentences} & \textbf{Single} & \textbf{Multiple} & \textbf{Avg.Sub-sents} & \textbf{Simile} \\
        \midrule[1.5pt]
        Training & 24697 & 90\% & 10\% & 6.4 & 70\% \\
        Validation & 500 & 99\% & 1\% & 2.9 & 90\% \\
        Test & 1001 & 99\% & 1\% & 2.9 & 90\% \\
        \bottomrule[1.5pt]
    \end{tabular}
\end{table*}

\subsection{Baselines}
To verify the effectiveness of our method, we compared it with the following baselines:

\textbf{LLM-Zero-Shot} Zero-shot experiments are conducted for comparison purposes on the recently released Chinese Tiny LLM \cite{du2024chinese}, MAP-Neo \cite{zhang2024map}, Qwen2-7B-Instruct \cite{qwen2}, Yi-1.5-34B\footnote{\url{https://huggingface.co/01-ai/Yi-1.5-34B}} and GPT-4-Turbo. The experimental results from NLPCC2024 Shared Task 9 are marked with $*$.

\textbf{LLM-Fixed} The method combines in-context learning with LLM fine-tuning, but the in-context examples are three fixed examples manually selected.

\textbf{LLM-Random} Similarly, in-context learning and LLM fine-tuning are incorporated. However, the in-context examples are randomly retrieved from the training set, rather than fixed examples.

\textbf{LLM-BERT} Analogous to the preceding two baselines, the distinction resides in the approach to example retrieval. Specifically, BERT \cite{devlin2018bert} encodes each sentence, subsequently leveraging the Euclidean distance within the vector space to quantify sentence similarity. The three most similar examples are then identified and incorporated into the fine-tuning instructions.

\textbf{LLM-Mix-tuning} It elegantly intertwines multi-task learning with LLM fine-tuning. In this case, multi-task learning is implemented by using mixed instruction fine-tuning, which first identifies metaphor components from data, and then selects the most suitable triplet option.

\subsection{Implementation Details}
Our framework, LaiDA, is trained in multiple stages. During the training of the GAT encoder, bert-base-chinese\footnote{\url{https://huggingface.co/google-bert/bert-base-chinese}} is used as the encoding model, and the top 20\% of similarity scores are regarded as positive samples and set to 1. The batch size is set to 256, and the initial learning rate is set to $1\times10^{-4}$. In the other stages of fine-tuning, Qwen2-7B-Instruct\footnote{\url{https://huggingface.co/Qwen/Qwen2-7B-Instruct}} is uniformly used as the backbone model, with low-rank adaptation (LoRA) \cite{hulora} employed for efficient parameter tuning. LoRA Rank is set to 32, batch size to 2, and the initial learning rate to 8e-5, with bf16 precision for training. All training is performed on a single 24GB NVIDIA GeForce RTX 4090 GPU, with accuracy as the primary evaluation metric.

\begin{table*}[t!]
    \caption{The overall performance of all the compared baselines and our LaiDA.}
    \label{tab:tab2}
    \centering
    \begin{tabular}{l|c|c}
        \toprule[1.5pt]
        \multicolumn{1}{c|}{\textbf{Methods}} & \textbf{Models} & \textbf{Acc.(\%)} \\
        \midrule[1.5pt]
        \multirow{5}{*}{LLM-Zero-shot} & Chinese Tiny LLM (2B) & 15.7* \\
        & MAP-NEO (7B) & 39.25* \\
        & Qwen2-Instruct (7B) & 71.43 \\
        & Yi-1.5 (34B) & 88* \\
        & GPT-4-Turbo & 89* \\
        \midrule[1pt]
        LLM-Fixed & Qwen2-Instruct (7B) & 92.21 \\
        LLM-Random & Qwen2-Instruct (7B) & 92.11 \\
        LLM-BERT & Qwen2-Instruct (7B) & 91.51 \\
        LLM-Mix-tuning & Qwen2-Instruct (7B) & 92.81 \\
        \midrule[1pt]
        LaiDA(ours) & Qwen2-Instruct (7B) & \textbf{93.21} \\
        \midrule[1pt]
        w/o Data Augmentation & Qwen2-Instruct (7B) & 92.31 \\
        w/o In-context Learning & Qwen2-Instruct (7B) & 90.61 \\
        \bottomrule[1.5pt]
    \end{tabular}
\end{table*}

\section{Results and Discussions}
We delve into the intricacies of our findings, commencing with an analysis of the main results, and individual metaphor component results. Furthermore, we conduct an ablation study to assess the contribution of each module in LaiDA and perform an error analysis to reveal potential avenues for improvement.

\textbf{Main Results} The overall performance of all the compared baselines and LaiDA is compared on the test set and reported in Table~\ref{tab:tab2}. LaiDA achieves the highest accuracy compared to several baselines, demonstrating its effectiveness. Interestingly, the application of pre-trained BERT \cite{devlin2018bert} for Euclidean distance similarity retrieval did not outperform random retrieval. The emphasis on similarity and diversity in selecting in-context examples might explain LLM-BERT's underwhelming performance versus LLM-Random, potentially stemming from Euclidean distance's inadequacy for similarity assessment and the broader diversity inherent in LLM-Random. Notably, LLM-Mix-tuning, leveraging multi-task learning, exhibits commendable results, partially attributed to the metaphor analysis prior to component selection, which augments the model. Nevertheless, despite doubling the training data, the enhancement is modest.

\textbf{Individual Metaphor Component Results} As shown in Table~\ref{tab:tab3}, LaiDA achieves an accuracy above 97\% in identifying the Tenor (97.20\%) and Vehicle (97.32\%), demonstrating its excellent capability in capturing the core elements of metaphors, attributed to their directness and clarity, which facilitate model detection. However, the accuracy in identifying the Ground (94.14\%) is slightly lower, indicating its higher difficulty. The Ground, serving as the bridge in metaphors, requires stronger contextual understanding and reasoning abilities from the model due to its subtle and abstract nature, highlighting it as a key area for future optimization.

\textbf{Ablation Study} To investigate the role of data augmentation and in-context learning in LaiDA, this section presents the results of ablation experiments. As shown in Table~\ref{tab:tab2}, removing the data augmentation pre-training module results in a 0.9\% decrease in accuracy. Similarly, removing in-context learning by not embedding any examples in the prompt during supervised fine-tuning leads to a 2.6\% decrease. These results confirm the effectiveness of both data augmentation pre-training and in-context learning in LaiDA.

\textbf{Error Analysis} Despite LaiDA's remarkable 93.21\% accuracy in MCI, a detailed error analysis is undertaken to bolster its performance, as presented in Table~\ref{tab:tab4}. It reveals that the only ground errors are as high as 35.94\%, indicating their obscure and variable nature within contexts, consistent with the phenomena reflected in Table~\ref{tab:tab3}. Notably, in samples with tenor or vehicle errors (including T, V, T-G, V-G, T-V-G), ground errors reach 87.89\%. More specifically, ground errors account for 82.15\% in tenor error samples, while this escalates to 92.61\% in vehicle errors, suggesting that tenor or vehicle misidentifications exacerbate ground identification difficulties, with vehicle errors exerting a more profound influence. Grounds, as connectors between tenor and vehicle, rely heavily on their accurate identification. Therefore, LaiDA's future advancements should prioritize enhancing contextual comprehension and refining tenor/vehicle detection for improved ground identification.

\begin{table}[!t]
    \centering
    \begin{minipage}[t]{0.37\textwidth}
        \caption{Performance of LaiDA on individual metaphor component.}
    \label{tab:tab3}
    \centering
    \begin{tabular}{c|c|c|c}
        \toprule[1.5pt]
        \textbf{Component} & \textbf{T} & \textbf{V} & \textbf{G} \\
        \midrule[1.5pt]
        \textbf{Acc.(\%)} & 97.20 & 97.32 & 94.14 \\
        \bottomrule[1.5pt]
    \end{tabular}
    \end{minipage}
    \hfill
    \begin{minipage}[t]{0.56\textwidth}
        \caption{Error analysis in LaiDA's results. The error types are specifically combined into seven types.}
    \label{tab:tab4}
    \raggedleft
    \begin{tabular}{c|c|c|c|c|c|c|c}
        \toprule[1.5pt]
        \textbf{Type} & \textbf{T} & \textbf{V} & \textbf{G} & \textbf{T-V} & \textbf{T-G} & \textbf{V-G} & \textbf{T-V-G} \\
        \midrule[1.5pt]
        \textbf{\%} & 4.69 & 0.00 & 35.94 & 3.12 & 17.19 & 20.31 & 18.75 \\
        \bottomrule[1.5pt]
    \end{tabular}
    \end{minipage}
\end{table}

\section{Conclusion}
 In this paper, a framework (LaiDA) integrating linguistics-aware in-context learning and data augmentation is constructed to enhance the accuracy and efficiency of MCI. A few raw data samples are initially processed using ChatGPT's superior semantic parsing capabilities to construct a benchmark dataset with incorrectly identified metaphor component triplet options. Supervised fine-tuning is then performed on this refined dataset to process the remaining raw data, resulting in a high-quality dataset tailored for training. Additionally, a simile dataset is used for pre-training to achieve data augmentation, allowing the model to grasp simpler metaphorical patterns in advance. A GAT encoder is introduced to efficiently retrieve similar contextual examples by capturing and encoding rich linguistic features into fine-grained representations, which are used to retrieve examples similar to the input query. Finally, we fine-tune the LLM on the constructed dataset using prompts that incorporate linguistically similar examples. The experimental results and analysis demonstrated the effectiveness of LaiDA in MCI.

\begin{credits}
\subsubsection{\discintname}
The authors have no competing interests to declare that are
relevant to the content of this article.
\end{credits}

\bibliographystyle{splncs04}
\bibliography{arxiv}
%
% \begin{thebibliography}{8}
% \bibitem{ref_article1}
% Author, F.: Article title. Journal \textbf{2}(5), 99--110 (2016)

% \bibitem{ref_lncs1}
% Author, F., Author, S.: Title of a proceedings paper. In: Editor,
% F., Editor, S. (eds.) CONFERENCE 2016, LNCS, vol. 9999, pp. 1--13.
% Springer, Heidelberg (2016). \doi{10.10007/1234567890}

% \bibitem{ref_book1}
% Author, F., Author, S., Author, T.: Book title. 2nd edn. Publisher,
% Location (1999)

% \bibitem{ref_proc1}
% Author, A.-B.: Contribution title. In: 9th International Proceedings
% on Proceedings, pp. 1--2. Publisher, Location (2010)

% \bibitem{ref_url1}
% LNCS Homepage, \url{http://www.springer.com/lncs}, last accessed 2023/10/25
% \end{thebibliography}
\end{document}